# A Comprehensive Performance Evaluation for 3D Transformation Estimation Techniques

Bao Zhao, Xiaobo Chen, Xinyi Le and Juntong Xi

**Abstract**—3D local feature extraction and matching is the basis for solving many tasks in the area of computer vision, such as 3D registration, modeling, recognition and retrieval. However, this process commonly draws into false correspondences, due to noise, limited features, occlusion, incomplete surface and etc. In order to estimate accurate transformation based on these corrupted correspondences, numerous transformation estimation techniques have been proposed. However, the merits, demerits and appropriate application for these methods are unclear owing to that no comprehensive evaluation for the performance of these methods has been conducted. This paper evaluates eleven state-of-the-art transformation estimation proposals on both descriptor based and synthetic correspondences. On descriptor based correspondences, several evaluation items (including the performance on different datasets, robustness to different overlap ratios and the performance of these technique combined with Iterative Closest Point (ICP), different local features and LRF/A techniques) of these methods are tested on four popular datasets acquired with different devices. On synthetic correspondences, the robustness of these methods to varying percentages of correct correspondences (PCC) is evaluated. In addition, we also evaluate the efficiencies of these methods. Finally, the merits, demerits and application guidance of these tested transformation estimation methods are summarized.

---✦---

## 1 INTRODUCTION

Local feature description is a fundamental and significant way for solving many tasks in computer vision area, such as 3D registration [1, 2], 3D object categorization and recognition [3-6], and 3D model retrieval and shape analysis [7, 8], to name a few. With the flourishment of low-cost sensors and high-performance computing systems, the significance of local feature descriptors is further improved in practice. In the last few years, a large number of local feature descriptors have been proposed, e.g., rotational projection statistics (RoPS) [9], Tri-Spin-Image (TriSI) [10] and multi-attribute statistics histograms (MaSH) [11]. For more detail, readers can refer to two surveys [12, 13].

In local feature descriptor based applications (e.g., 3D registration [1, 14] and object recognition [3, 4, 9]), keypoints are first extracted. On these keypoints, local features are generated. Then, the local features on one surface are matched with the local features on the other surface to obtain point-to-point correspondences. Finally, a correct transformation is estimated from the constructed correspondences for registering the two surfaces. In the above process, the correctness of an estimated transformation is critical for the success of aligning the two surfaces. However, the generated correspondences may contain a large number of outliers due to symmetric structures, noise, clutter and occlusions. Estimating a correct transformation on the seriously corrupted correspondences is a big challenge. To address this issue, extensive transformation estimation techniques have been proposed, e.g., the random sample consensus (RANSAC) [15], sample consensus initial alignment (SAC-IA) [16], Game theoretical matching [17, 18], Regularization for Iterative Re-Weighting (RIRW) [19]. Here, we broadly classify these techniques into two categories: *maximum consistency* (MC)-based and *confidence verify* (CV)-based methods. MC-based methods attempt to acquire correct correspondences and eliminate false ones as many as possible from the corrupted correspondences, as the pipeline (a)-(c)-(d) shown in Fig. 1. CV-based methods commonly first use a particular way for estimating a plausible transformation, and then verify the confidence level of this transformation by using it to align two point clouds, as the pipeline (a)-(b)-(d) shown in Fig. 1. In CV-based methods, the way of estimating a transformation mainly includes three methods: using one correspondence and associated Local Reference Frame (LRF) which comprises three orthogonal axes, using two correspondences and associated Local Reference Axis (LRA) which only contains a single orientated axis, employing three correspondences or more. For avoiding redundancy, the detailed description of the state-of-the-art transformation estimation methods is presented in Section 2.

The performance evaluations or surveys for the techniques involved in the pipeline of 3D registration and object recognition have been widely proposed, e.g., local feature descriptors [12, 13], keypoint detectors [20], LRF and LRA [21] and 3D registration [22-24]. To the best of our knowledge, no survey article has been proposed for comprehensively evaluating the performance of the state-of-the-art transformation estimation techniques, though

---

- *B. Zhao, X. Chen, X. Le and J. Xi are with the School of Mechanical Engineering, Shanghai Jiao Tong University, Shanghai, 200240, China. E-mails {zhaobao1988; xiaoboc; lexinyi; jtxi}@sjtu.edu.cn.*
- *Corresponding author: J. Xi.*

these techniques are very critical in the process of 3D registration and object recognition. As a result, the merits, demerits and application scenarios of these transformation estimation methods are unclear so far.

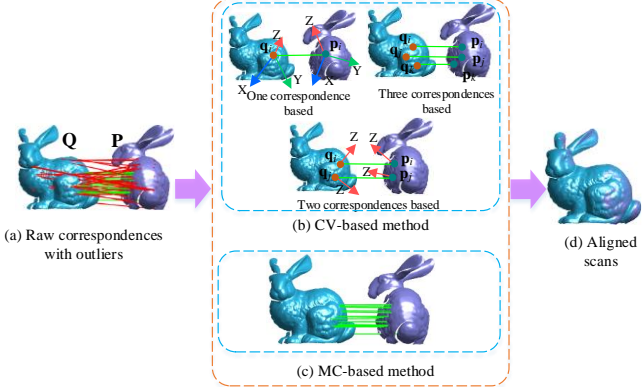

Fig. 1. An illustration for the pipeline of transformation estimation. (a) The original correspondences with outliers generated by local feature descriptors. (b) Transformation estimated by CV-based method. (c) Transformation estimated by MC-based method. (d) Aligned scans by using estimated transformation.

Given the above considerations, we conduct an extensive survey of the state-of-the-art transformation estimation methods by first presenting the major technique steps in these methods and then implementing comprehensive experimental evaluation. Specifically, eleven methods are considered in this paper, and they are tested on four popular datasets with different data modalities (e.g., Minolta vivid, Kinect and Space Time). For an unbiased evaluation, some important and common parameters for the eleven methods are tested. Based on the parameter settings, the performance of the eleven methods are first evaluated on the correspondences generated by local descriptor. In this process, several items (including the performance on different datasets, robustness to different overlap ratios and the performance of these techniques combined with ICP algorithm [25], different local features and LRF/A techniques) are tested. In addition, the performance on synthetic correspondences and the efficiency of these methods are also evaluated. Two main contributions of this paper are summarized as follows:

(1) We provide a comprehensive review and quantitative evaluation of eleven state-of-the-art algorithms on four benchmarks acquired by some common devices (e.g., Minolta vivid, Kinect and Space Time). Various relevant items with transformation estimation techniques are tested. To the best of our knowledge, this is the first survey paper that comprehensively evaluates the transformation estimation techniques in 3D space.

(2) The merits and demerits of each method are summarized and discussed, giving instructive information for developers to select an appropriate technique in their particular applications.

The rest of this paper is organized as follows. Section 2 presents the main technique steps of eleven state-of-the-art algorithms. The evaluation methodology is detailed in Section 3, and the experimental results are drawn in Section 4. Summary and analysis of experimental results are shown in Section 5, and the conclusion is reported in Section 6.

## 2 CONSIDERED METHODS

Existing transformation estimation techniques, as illustrated in Fig. 1, can be broadly categorized into two groups: *maximum consistency* (MC) based methods and *confidence verification* (CV) based methods. The CV-based methods are defined as the methods which need to verify the reliability of a plausible transformation in each iteration. Commonly, the reliability is verified by the accuracy of two aligned scans using the plausible transformation. The *MC*-based methods are defined as the methods which first search the maximum consistency set from initial correspondences, and then estimate a transformation based on the obtained consistency set. Since the MC-based methods do not need to verify the confidence for a plausible transformation, they usually present a high efficiency. In this paper, six MC-based methods and four CV-based methods are considered. In addition, the Consistent Correspondences Verification (CCV) method is also considered, which have the features of both MC-based CV-based methods. For readability, CCV technique is classified as a CV-based method. Some common notations used in describing these methods are reported in Table 1. In addition, in the process of registering two scans, we treat the transformed scan as model and the fixed scan as scene for the simplicity of this paper.

TABLE 1
Notations Used in This Paper.

| Notations | Definitions |
|---|---|
| $\mathbf{P}$ and $\mathbf{Q}$. | A model and scene, respectively. |
| $\mathbf{P}_k=\{\mathbf{p}_1, \mathbf{p}_2,…, \mathbf{p}_m\}$ and $\mathbf{Q}_k=\{\mathbf{q}_1, \mathbf{q}_2,…, \mathbf{q}_n\}$ | The keypoints selected on $\mathbf{P}$ and $\mathbf{Q}$, respectively. |
| $\mathbf{C}=\{\mathbf{c}_1, \mathbf{c}_2,…, \mathbf{c}_s\}$ | A set of correspondences. |
| $c_i(\mathbf{p}_i, \mathbf{q}_i)$ | A correspondence composed of $\mathbf{p}_i$ and $\mathbf{q}_i$. |
| $d(\mathbf{p}_i, \mathbf{p}_j)$ | The Euclidean distance between $\mathbf{p}_i$ and $\mathbf{p}_j$. |
| LRF($\mathbf{p}_i$) and LRA($\mathbf{p}_i$) | The LRF and LRA at $\mathbf{p}_i$, respectively. |

### 2.1 MC-based Techniques

#### 2.1.1. Random Sample Consensus (RANSAC) [15]

RANSAC is a well-known algorithm widely used in the area of computer vision. The core of this algorithm is to iteratively obtain the parameters with the maximum consensus set. In the application of rejecting false correspondences in 3D case, three correspondences at least are randomly selected in each iteration for determining a transformation. Specifically, three steps are iteratively implemented. *First*, three correspondences at least are sampled from all given correspondences. *Second*, a plausible transformation is calculated by the selected correspondences, and then the model keypoints in all given correspondences are transformed to align with the corresponded scene keypoints using the obtained transformation. Then, the correspondences with the distance errors below a threshold $t$ are served as a consensus set. *Third*, if the number of correspondences in the consensus set is larger than a pre-defined threshold $k$, a transformation is calculated based on the obtained consensus set, and if not, the algorithm directly goes to the first step and repeat this process. The above three steps are repeated until the number of iterations reaches a predetermined value. If the iteration is end, the transformation corre-sponding to the consensus set with a largest number of correspondences is selected as the output transformation.

### 2.1.2. Geometric Constraint Cluster (GCC) [26]

In this method, a geometric constraint in rigid object is used, which is denoted as:

$$\left|d(\mathbf{q}_i, \mathbf{q}_j) - d(\mathbf{p}_i, \mathbf{p}_j)\right| < \varepsilon, \quad (1)$$

where $\varepsilon$ is a pre-defined threshold. This method mainly includes three steps: *First*, each correspondence $c_i(\mathbf{p}_i, \mathbf{q}_i)$ is used as an initial matched pair in one group; *Second*, for each group, all other correspondences are added in it if they satisfy Eq (1); *Third*, the group with the largest size of correspondences is selected to calculate the optimal transformation.

### 2.1.3. Geometric Constraint-based Method (GCM) [27]

This method also employs the rigidity constraint calculated by Eq (1) to exclude some correspondences with inferior consistency confidence. Specifically, each correspondence is evaluated with all other correspondences in **C**. If the number of correspondence pairs violating the rigidity constraint exceeds a certain percentage (denoted as $\delta$), the current correspondence is deemed as an incorrect correspondence, and then removed from **C**. Finally, the correspondences retained in **C** are used to calculate the output transformation. In addition, for further handling the case with the majority of correspondences being false, dynamically updating the thresholds (i.e., $\varepsilon$ and $\delta$) is implemented in the above process. In particular, the thresholds are loosely set in the begin of this algorithm, and the thresholds are severely set in the end of this algorithm. For accurately evaluating the performance of this method, the dynamically thresholds are used in our evaluation process. For simplicity, the loose and severe thresholds are denoted as $(\varepsilon_1, \delta_1)$ and $(\varepsilon_2, \delta_2)$, respectively.

### 2.1.4. Game Theoretic Matching (GTM) [18, 28]

This method firstly uses the concept of game theory in 3D matching purpose. Similar to a game $G$, each point cloud (**P** or **Q**) is a player and all given correspondences **C** form a set of pure strategies. The game theory is summarized as a triplet $G= \{\mathbf{I}, \mathbf{C}, \pi\}$, where **I** is the player set $\{\mathbf{P}, \mathbf{Q}\}$ and $\pi$ is a combined payoff function. At equilibrium, only the mutually compatible correspondences are preserved and considered as inliers, and these incompatible correspondences are dismissed.

In this framework, a key step is to construct the payoff matrix $\mathbf{\pi}$. The payoff matrix, usually denoted by a real valued function $\pi: \mathbf{C} \times \mathbf{C} \rightarrow R^+$, is materialized in a symmetric payoff matrix $\mathbf{\pi}$ to quantify the degree of compatibility between each correspondence and the others. In the application of estimating transformation for rigid objects, the payoff function $\mathbf{\pi}$, defined as Eq (2), is constructed as a geometric constraint to evaluate the compatibility between any two correspondence pairs.

$$\pi(c_i(\mathbf{p}_i, \mathbf{q}_i), c_j(\mathbf{p}_j, \mathbf{q}_j)) = \left(\frac{\min(d(\mathbf{p}_i, \mathbf{p}_j), d(\mathbf{q}_i, \mathbf{q}_j))}{\max(d(\mathbf{p}_i, \mathbf{p}_j), d(\mathbf{q}_i, \mathbf{q}_j))}\right)^\lambda, \quad (2)$$

where $\lambda$ is a pre-defined parameter. After calculating the payoff values between any two correspondence pairs, the symmetric payoff matrix $\mathbf{\pi}$ can be materialized. Then, a probability distribution $x \in \Delta^{|\mathbf{C}|} = \{x \in R^{|\mathbf{C}|} : \sum_{i=1}^{|\mathbf{C}|} x_i = 1 \text{ and } x_i \geq 0\}$ over the strategy set **C** can be iteratively evolved by applying the replicator dynamics equation:

$$x_i(k+1) = x_i(k) \frac{(\mathbf{\pi} x(k))_i}{x(k)^T \mathbf{\pi} x(k)}, \quad (3)$$

where $k$ is the number of iteration and $\mathbf{\pi}$ is the payoff matrix. The initial values of each element in $x$ can be set to $1/|\mathbf{C}|$. When $x$ converges, a correspondence $c_i \in \mathbf{C}$ is considered as true if $x_i >= t$, and it is deemed as false if $x_i < t$, where $t$ is a pre-defined parameter. Once a set of plausible correspondences are obtained, the rigid transformation can be easy calculated by some close form techniques [29-31].

### 2.1.5. A Variant of GTM (V-GTM) [32]

This method is a variant of the GTM technique introduced in Section 2.1.4. In contrast to the GTM technique, this method adds a geometric constraint with an exponential form into the payoff function Eq (2). The purpose of adding this geometric constraint is to reduce the compatibility between two correspondence pairs with large distance deviation. The upgraded payoff function is calculated as:

$$\pi(c_i(\mathbf{p}_i, \mathbf{q}_i), c_j(\mathbf{p}_j, \mathbf{q}_j)) = \frac{\min(d(\mathbf{p}_i, \mathbf{p}_j), d(\mathbf{q}_i, \mathbf{q}_j))}{\max(d(\mathbf{p}_i, \mathbf{p}_j), d(\mathbf{q}_i, \mathbf{q}_j))} e^{-\frac{|d(\mathbf{p}_i, \mathbf{p}_j) - d(\mathbf{q}_i, \mathbf{q}_j)|}{\gamma}}, \quad (4)$$

where $|\ |$ denotes the operator of taking absolute value. In addition, for improving efficiency and restricting the transformation solution with one-to-one, the final payoff function $\mathbf{\pi}$ is defined as follows:

$$\pi(c_i, c_j) = \begin{cases} 0, & \text{if } \mathbf{p}_i = \mathbf{p}_j \text{ or } \mathbf{q}_i = \mathbf{q}_j \text{ or } \pi(c_i, c_j) < 0.1 \\ \pi(c_i, c_j), & \text{otherwise} \end{cases} \quad (5)$$

After defining the payoff function, the initialization and evolution of the probability distribution $x$ are performed with the same techniques of the GTM method as presented in Section 2.1.4. When the evolution is terminated, a correspondence $c_i \in \mathbf{C}$ is considered as true if $x_i > 0$, and it is deemed as false if $x_i = 0$. On this basis, we can find that, in contrast to the GTM method, the other advantage of this method is that it does not need to set a threshold for excluding false correspondences. Once a set of plausible correspondences are obtained, the rigid transformation can be easy calculated by some close form techniques [29-31].

### 2.1.6. Local and Global Voting (LGV) [33]

This method is implemented in two steps: local and global voting stages. In local voting stage, for a correspondence $c_i(\mathbf{p}_i, \mathbf{q}_i)$, a set of neighbors $N(c_i)$ is first searched. The compatibility, denoted as $u_L(c_i, c_j)$, between $c_i$ and each correspondence in $N(c_i)$ is measured by also using the Eq (2) with the $\lambda$ set to 1. Next, a subset of correspondences with high compatibility to $c_i$ is extracted as:

$$\gamma_L(c_i) = \{c_j \in N(c_i) : u_L(c_i, c_j) > \zeta\}, \quad (6)$$

where $\zeta \in [0,1]$ is a pre-defined threshold. Then, a local score, denoted as $S_L$, for evaluating the evidence of the correspondence $c_i$ in the local constraint is measured by:

$$S_L(c_i) = \frac{|\gamma_L(c_i)|}{|N_L(c_i)|}, \tag{7}$$

where | | is an operator for extracting the number of elements in a set.

In global voting stage, the top $k$ correspondences, denoted as $\mathbf{C}_G$, with high $S_L$ values are first selected. In this stage, the LRF associated with all correspondences in $\mathbf{C}_G$ are needed to be available. On this basis, a plausible transformation to a correspondence $c_i(\mathbf{p}_i, \mathbf{q}_i)$ can be calculated as:

$$\begin{cases} \mathbf{R}_i = (\text{LRF}(\mathbf{q}_i))^\text{T} \text{LRF}(\mathbf{p}_i) \\ \mathbf{t}_i = \mathbf{q}_i - \mathbf{R}_i \mathbf{p}_i \end{cases}, \tag{8}$$

where $\mathbf{R}_i$ and $\mathbf{t}_i$ denote the rotation matrix and translation vector, respectively. By using this transformation, the model points in $\mathbf{C}_G$ can be transformed to align with the corresponded scene points. Then, a global confidence, denoted as $u_G(c_i, c_j)$, can be evaluated by:

$$u_G(c_i, c_j) := d(\mathbf{T}(c_i)\mathbf{p}_j, \mathbf{q}_j), \tag{9}$$

where $\mathbf{T}(c_i)=(\mathbf{R}_i, \mathbf{T}_i)$ denotes the transformation corresponding to $c_i$. Then, a global vote $\gamma_G(c_i)$ is calculated as:

$$\gamma_G(c_i) = \{c_j \in \mathbf{C}_G : u_L(c_i, c_j) > \zeta \land u_G(c_i, c_j) < \delta\}, \tag{10}$$

where $\delta$ is a pre-defined distance tolerance. The final score $S(c_i)$ is calculated by combining the local and global votes as:

$$S(c_i) = \frac{|\gamma_L(c_i)| + |\gamma_G(c_i)|}{|N_L(c_i)| + |\mathbf{C}_G(c_i)|} \tag{11}$$

Finally, the Otsu's adaptive threshold method [34] is used to separate the inliers from $\mathbf{C}_G$ based on the final score $S$. Based on the obtained inliers, the transformation can be easy calculated by some close form techniques [29-31].

## 2.2 CV-based Techniques

### 2.2.1. Sample Consensus Initial Alignment (SAC-IA) [16]

This algorithm is developed from the RANSAC method for iteratively performing two steps. *First*, three correspondences at least are sampled from all given correspondences $\mathbf{C}$. In this process, the distances between two sampled model points are ensured to larger than a minimum distance. *Second*, a rigid transformation is calculated by the sampled correspondences and evaluated by an error metric with a Huber penalty measure $L_h$:

$$L_h(e_i) = \begin{cases} \frac{1}{2} e_i^2 & \|e_i\| \leq t_e \\ \frac{1}{2} t_e (2\|e_i\| - t_e) & \|e_i\| > t_e \end{cases}, \tag{12}$$

where $e_i$ is a distance error between one transformed model point and its closet scene point. If the number of iteration satisfies a pre-defined maximum count, the iteration is terminated. If the iteration is end, a plausible transformation corresponding to a minimum error metric is selected as the optimal transformation.

### 2.2.2. Consistent Correspondences Verification (CCV) [9]

This algorithm is mainly composed of four steps. *First*, a plausible transformation $\mathbf{T}(c_i)=(\mathbf{R}_i, \mathbf{T}_i)$ for each correspondence $c_i$ is first calculated by using the Eq (8). Note that the LRF associated with all correspondences $\mathbf{C}$ are also needed to be available in this method. *Second*, a consistent set associated with each correspondence is obtained by clustering algorithm. Specifically, the rotation matrix in each estimated transformation is converted into three Euler angles for effectively implementing clustering calculation. Then, the Euclidean distances of Euler angles and translation vectors between any two transformations are measured. For each estimated transformation $\mathbf{T}(c_i)$, the remaining transformations with the angle distances to $\mathbf{T}(c_i)$ less than $\tau_a$ and the translation distances to $\mathbf{T}(c_i)$ less than $\tau_t$ are grouped into a cluster and viewed as a consistent set, where the $\tau_a$ and $\tau_t$ are pre-defined thresholds. *Third*, the plausible transformation associated with each correspondence is calculated by all correspondences in its consistent set. *Fourth*, in order to obtain a more accurate transformation, these plausible transformations are verified by the statistic of the number of inliers. In particular, the model $\mathbf{P}$ and scene $\mathbf{Q}$ are first randomly simplified. For each plausible transformation, the simplified $\mathbf{P}$ is transformed to align with the simplified $\mathbf{Q}$. Then, each point in the simplified $\mathbf{P}$ is corresponded with the closest point searched on the simplified $\mathbf{Q}$. The number of inliers, which defined as the correspondences with the distance errors less than two times of average mesh resolution (mr), is counted. Finally, the plausible transformation corresponded to the maximum number of inliers is selected as the final transformation estimation. It is worth noting that using the number of inliers for verifying the confidence of a plausible transformation is also employed in 1-Point random sample consensus (1P-RANSAC) [35], Optimized sample consensus (OSAC) [1] and 2-point based sample consensus with global constrain (2SAC-GC) [11].

### 2.2.3. 1-Point Random Sample Consensus (1P-RANSAC) [35]

This algorithm is conducted also based on the LRF transformation technique as presented in the Eq (8). Specifically, a correspondence is randomly selected in each iteration for calculating a plausible transformation using the Eq (8). Then, the confidence of the plausible transformation is verified by counting the number of inliers as detailed in Section 2.2.2. If the size of inliers in current iteration larger than that in all previous iteration, the optimal transformation is updated with the transformation calculated by all the inliers obtained in this iteration. The above iteration process continues until the maximum iteration count is reached. After the iteration is terminated, the finally optimal transformation is obtained.

### 2.2.4. Optimized Sample Consensus (OSAC) [1]

In this algorithm, two steps are iteratively implemented. *First*, three correspondences at least are sampled from all given correspondences. In this process, the distances between two sampled points on model are ensured to larger than a minimum distance. *Second*, a rigid transformation is calculated by the sampled correspondences and evaluated by an error metric defined as:

$$D_{avg}(\tilde{\mathbf{P}}, \tilde{\mathbf{Q}}) = \begin{cases} \frac{1}{\hat{N}} \sum_{i=1}^{\hat{N}} d(\tilde{\mathbf{p}}_i, \tilde{\mathbf{q}}_i) & \text{if } \frac{\hat{N}}{\min\{N_1, N_2\}} > \delta \\ \infty & \text{otherwise} \end{cases}, \quad (13)$$

where $\tilde{\mathbf{P}}$ and $\tilde{\mathbf{Q}}$ denote the simplified $\mathbf{P}$ and $\mathbf{Q}$, respectively; $N_1$ and $N_2$ denote the number of points in $\tilde{\mathbf{P}}$ and $\tilde{\mathbf{Q}}$, respectively; $\hat{N}$ denotes the number of inliers and $\delta$ is a pre-defined threshold used to judge if $\tilde{\mathbf{P}}$ and $\tilde{\mathbf{Q}}$ are spatially close. If the number of iteration satisfies a user-defined maximum count, the iteration is terminated. Then, the transformation corresponded to the minimum error metric is selected as the output transformation.

### 2.2.5. 2-point Based Sample Consensus with Global Constrain (2SAC-GC) [11]

In contrast to CCV and 1-P RANSAC algorithms of using one correspondence to determine a transformation, 2SAC-GC randomly selects two correspondences to calculate a plausible transformation based on LRA. The reason of using LRA rather than LRF to calculate a transformation is that an LRA commonly presents higher repeatability than an LRF [36, 37]. In this algorithm, two point correspondences $c_i(\mathbf{p}_i, \mathbf{q}_i)$ and $c_j(\mathbf{p}_j, \mathbf{q}_j)$ are randomly selected in each iteration. The two selected correspondences are first evaluated by two constraints as:

$$\begin{cases} |d(\mathbf{p}_i, \mathbf{p}_j) - d(\mathbf{q}_i, \mathbf{q}_j)| \le \sigma_d \\ |\arccos(\text{LRA}(\mathbf{p}_i) \cdot \text{LRA}(\mathbf{p}_j)) - \\ \quad \arccos(\text{LRA}(\mathbf{q}_i) \cdot \text{LRA}(\mathbf{q}_j))| < \sigma_a \end{cases}, \quad (14)$$

where $|\ |$ represents an operator to take absolute value, and $\sigma_d$ and $\sigma_a$ denote the pre-defined distance and angle thresholds, respectively. If the two selected correspondences satisfy the above two constraints, a plausible transformation is calculated by them. Otherwise, the algorithm directly switches to the next iteration. After obtaining a plausible transformation, the size of the inliers is counted by the same way in Section 2.2.2. If the number of iteration satisfies a per-defined maximum value, this iteration process is determined, and the plausible transformation corresponded to the maximum number of inliers is selected as the output transformation.

## 3 EVALUATION METHODOLOGY

### 3.1. Datasets and Evaluation Criteria

#### 3.1.1. Datasets

In this paper, all the eleven transformation estimation techniques are tested on four datasets, i.e., Stanford 3D Modeling (S3M) [20, 38], UWA 3D Modeling (U3M) [5, 39], Kinect 3D Registration (K3R) and Stereo 3D Registration (S3R) datasets. The examples of these datasets are presented in Fig. 2 and the features of these datasets are reported in Table 2. The main considerations for selecting these datasets are threefold. First, various acquisition devices are used. Specifically, S3M and U3M datasets are acquired by a Minolta vivid scanner. K3R and S3R datasets are acquired by a Microsoft Kinect and a Space Time Stereo, respectively. The point clouds acquired by the Minolta vivid scanner are dense and high quality, and the point clouds acquired by the Microsoft Kinect and the Space Time Stereo are sparse and noisy. These three acquisition devices are commonly used in practice, which ensures the applied worth of this evaluation. Second, various challenges are contained in these four datasets. In particular, self-occlusion and missing regions are contained in all the four datasets. Real noise is existed in the K3R and S3R datasets, and the outliers are existed in the S3R dataset. Since it is not true that all the scan pairs have overlaps, only the pairs whose overlap ratios are larger than 10% are used in this paper. The overlap ratio between any two scans ($p_s$ and $p_t$) is computed as:

$$\text{Overlap ratio} = \frac{\text{No. of correspondences in } p_s \text{ and } p_t}{\min(\text{No. of points in } p_s, \text{No. of points in } p_t)} \quad (15)$$

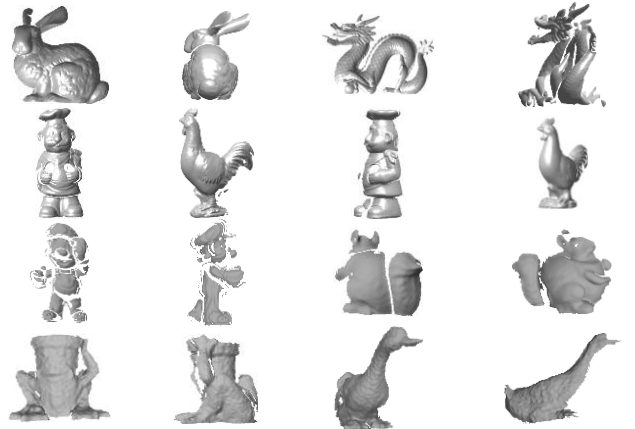

Fig. 2. Four exemplar scans respectively taken from each of the B3R, U3OR, U3M and QuLD datasets. From the first line to the fourth line are S3M, U3M, S3R and K3R datasets, respectively.

TABLE 2
Experimental Datasets and Their Features.

| Dataset | Challenge | Acquisition | Quality | # Number of scans | # Matching Pairs |
|---|---|---|---|---|---|
| Stanford 3D Modeling (S3M) | Self-occlusion and missing regions | Minolta vivid | high | 148 | 1382 |
| UWA 3D Modeling (U3M) | Self-occlusion and missing regions | Minolta vivid | high | 75 | 433 |
| Kinect 3D Registration (K3R) | Self-occlusion, missing regions and real noise | Microsoft kinect | low | 64 | 284 |
| Stereo 3D Registration (S3R) | Self-occlusion, missing regions, real noise and outliers | SpaceTime Stereo | low | 57 | 240 |

### 3.1.2. Evaluation Criteria

Two criteria are employed for evaluating the performance of all eleven tested transformation estimation techniques. The two criteria are the rotation error between the estimated rotation matrix $\mathbf{R}_E$ and the ground truth one $\mathbf{R}_{GT}$, and the translation error between the estimated translation vector $\mathbf{t}_E$ and the ground truth one $\mathbf{t}_{GT}$. The ground truth $\mathbf{R}_{GT}$ and $\mathbf{t}_{GT}$ for each valid scan pairs are known a prior, either provided by the publishers or obtained by manual alignment. Considering that the translation error is coupled with rotation movement, the influence of rotation movement need to be compensated when calculate the translation error. Assume that the $\mathbf{m}_C$ denotes the center of one scan needed to be transformed by $\mathbf{R}_E$ and $\mathbf{t}_E$. The compensation vector is denoted as: $\mathbf{R}_{GT}\mathbf{m}_C - \mathbf{R}_E \mathbf{m}_C$. On this basis, the rotation error $\varepsilon_r$ and the translation error $\varepsilon_t$ are defined as:

$$\begin{cases} \varepsilon_r = \arccos\left(\frac{trace(\mathbf{R}_{GT}(\mathbf{R}_E)^{-1})-1}{2}\right)\frac{180}{\pi} \\ \varepsilon_t = \frac{\|\mathbf{t}_{GT} - \mathbf{t}_E + \mathbf{R}_{GT}\mathbf{m}_C - \mathbf{R}_E \mathbf{m}_C\|}{mr} \end{cases}, \quad (16)$$

where mr denotes the average mesh resolution. For an estimated transformation, it is deemed as a correct transformation if $\varepsilon_r$ below 5° and $\varepsilon_t$ below 5, otherwise regarded as a false one.

### 3.2. Correspondences Generation

For descriptor based correspondences, local features need to be generated firstly. Specifically, we first randomly sample a small subset of points on original scan as key points. Note that, though a number of 3D keypoint detection techniques [40-43] have been proposed, most of them are time-consuming. Thus, random extraction is used in this paper. After obtaining the keypoints for each scan, the LRF technique proposed by Yang et al. [44], which is verified with the high robustness to mesh boundary [21], are employed for building the LRFs on the selected key points. Then, we use the Statistic of Deviation Angles on Subdivided Space (SDASS) descriptor [45] to generate local features on the selected key points. SDASS is a superior local descriptor as verified in [45]. It is worth mentioning that the LRA used in SDASS is replaced by the above built LRA for improving the robustness of local feature to mesh boundary. Note again that some transformation estimation techniques (e.g., 1P-RANSAC and 2SAC-GC) depending on LRF/A also use the above built LRF. After generating the local features, the most similar scene feature to each model feature is searched by kd-tree technique to obtain model-scene feature pairs, whose corresponding point pairs are taken as original correspondences. For improving the efficiency of registration, only some superior correspondences extracted from the original correspondences are used to perform transformation estimation. The detail of extracting some superior correspondences is introduced in Section 3.3.1.

Besides testing on descriptor based correspondences, the test on synthetic correspondences with different percentages of correct correspondences (PCC) is also implemented. In contrast to descriptor based correspondences, synthetic correspondences are not impacted by the nuisances contained in scans (e.g., symmetric surface, noise, and etc.), and only relevant to the value of PCC. Given the number of synthetic correspondences and a PCC value, the number of correct and false correspondences can be determined. In this experiment, correct correspondences are generated by first randomly selecting the points on the overlap of model and then extracting the corresponded points on scene based on ground truth transformation, and false correspondences are constructed by randomly selecting points on both model and scene. The number of synthetic correspondences for all transformation techniques are consistently set to 200, and the PCC values is set in Section 3.4.2.

### 3.3. Parameter Settings

The setting to some specific parameters of the evaluated methods (except for 1P-RANSAC, in which no specific parameters are needed to be set) are listed in Table 3. Most of these parameter settings are recommended by the original literatures, and the remaining parameter settings are set through a lot of experiments. In addition, some important and common parameters (e.g., the number of iteration) are tested in this paper for an unbiased evaluation. Specifically, three important parameters are tested in this paper. The first parameter is the number of inputting correspondences (tested in Section 3.3.1). The second parameter is the number of points sampled from model and scene to verify the confidence of an estimated transformation (tested in Section 3.3.2). The third parameter is the maximum number of iteration (tested in Section 3.3.3). The reasons of setting these three parameters are threefold. First, these parameters are commonly used in the tested transformation estimation techniques, e.g., the first parameter correlative with all the methods, the second parameter used in all the CV-based methods and the third parameter employed in SAC-IA, 1-P RANSAC, OSAC and 2SAC-GC. Second, these parameters are rarely tested before, especially for the first two parameters which have not been tested before to the best of my knowledge. Third, these parameters seriously affect the efficiency and accuracy of the algorithms. For simplicity, we only use U3M dataset to implement this test.

TABLE 3
The Settings to Some Specific Parameters of the Evaluated Transformation Estimation Methods.

| Methods | Parameters | Methods | Parameters |
|---|---|---|---|
| RANSAC | $t$=3mr, $k$=5 | CCV | $\tau_a$=0.2; $\tau_t$=10mr |
| GCC | $\varepsilon$=5mr | LGV | $\zeta$=0.9, $k$=250, $\delta$=5mr |
| SAC-IA | $t_e$=2mr | OSAC | $\delta$=0.3 |
| GTM | $\lambda$=1, $t$=0.5 | 2SAC-GC | $\sigma_d$ = 6mr, $\sigma_a$ =6° |
| GCM | $\varepsilon_1$=8mr, $\delta_1$=70%; $\varepsilon_2$=2mr, $\delta_2$=30% | V-GTM | $\gamma$=1 |

### 3.3.1. Setting the Number of Inputting Correspondences

Using all original correspondences to estimate a transformation is time-consuming as well as inaccurate. In practice, some superior correspondences are first extracted from the original ones before performing trans-formation estimation. So far, two approaches are commonly used for achieving this purpose. The first one, as used in [10, 46-48], is to search two closest scene features for each model feature, and then extract a number of correspondences with less ratio between the smallest distance and the second smallest one. The second one, as employed in [1, 11, 49], extract a number of most similar feature pairs. However, it is not tested before that which method is better and how many correspondences being extracted is appropriate. For simplicity, the above two techniques are denoted as *ratio* and *similarity* based methods, respectively. In this test, we first compare the performance of *ratio* and *similarity* based methods, and then select a superior method to test the appropriate number of input correspondences for these transformation estimation techniques. Besides, the number of input correspondences is set from 30 to 300 increased by 30 in this test.

The results of PCC acquired by *ratio* and *similarity* based methods are shown in Fig. 3. We can observe that *ratio* based method significantly outperforms *similarity* based method in terms of PCC, and the PCC performance of both *ratio* and *similarity* based methods gradually drop along with the increase of the number of input correspondences. It is clear that the *ratio* based method is obviously superior to the *similarity* based method. Thus, we consistently use the *ratio* based method to extract some superior correspondences hereinafter.

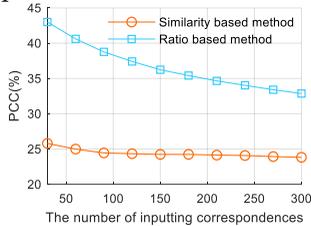

Fig. 3. The percentage of correct correspondences (PCC) with respect to different number of input correspondences acquired by ratio and similarity based methods.

Based on the *ratio* based method, the ratio of correct registration achieved by all evaluated transformation estimation techniques in terms of different number of input correspondences are tested. The results are presented in Fig. 4. We can find that GCC, GCM, 2SAC-GC and SAC-IA have superior performance under the number of input correspondences about 50. RANSAC and OSAC achieve superior performance under the number of inputting correspondences about 100. 1P-RANSAC exhibits superior performance under the number of inputting correspondences about 150. V-GTM, LGV, GTM and CCV present superior performance under the number of inputting correspondences about 200. Based on the above results and the trade-off between accuracy and efficiency, the number of input correspondences for RANSAC, GCC, SAC-IA, GTM, GCM, CCV, LGV, 1P-RANSAC, OSAC, 2SAC-GC and V-GTM are set to 100, 50, 50, 200, 50, 200, 200, 150, 100, 50 and 200, respectively.

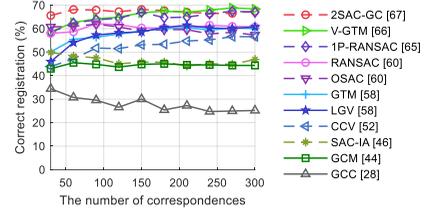

Fig. 4. The percentages of correct registration for all evaluated methods with respect to different number of input correspondences. The numbers in square brackets are the average percentages computed over the corresponding curves, and listed in a descending order.

### 3.3.2. Setting the Number of Points in Verifying Confidence

In the test for the second parameter, two groups of experiment are set. In the first group, we vary the number of points sampled on model and fix the number of points sampled on scene. In the second group, we vary the number of points sampled on scene and fix the number of points sampled on model to the best value selected from the first test. Specifically, in the first group, the number of points randomly sampled on model is set to 10, 50, 100, 500 and 1000, and the number of points randomly sampled on scene is fixed to 10000. In the second group, the number of points randomly sampled on scene is increased from 2000 to 10000 with an interval of 1000. Note that, since the resolution in a pairwise registration is usually ensured by the resolution of scene, the number of point in scene is set to a large value.

The results of setting the number of points used in verifying the confidence of an estimated transformation are shown in Fig. 5. It is worth noting that this parameter is only used in the CV-based methods (i.e., SAC-IA, CCV, 1P-RANSAC, OSAC and 2SAC-GC). In the test for setting sampled model points, the percentages of correct registration achieve an obvious promotion when the number of model points increased from 10 to 50, and then almost does not improve as the number of model points further increased. For a trade-off between the percentage of correct registration and efficiency, the number of points sampled on model is set to 100. In the test for setting sampled scene points, the number of model points is fixed to 100. The correct percentage improves in a slight manner as the scene points increased from 2000 to 8000, and then almost does not improve as the further increase of scene points. Based on this observation, the number of scene points is set to 8000.

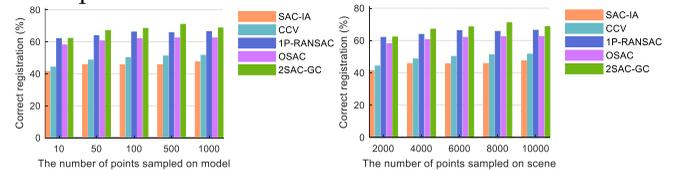

Fig. 5. The percentages of correct registration with respect to different number of points used for verifying the confidence to an estimated transformation. Left: testing the number of points sampled on model. Right: Testing the number of points sampled on scene.

### 3.3.3. Setting the Number of Iteration

In setting this parameter, five levels of this parameter, increased from 100 to 500 with an interval of 100, are tested. Note that this parameter is only employed in SAC-IA, 1P-RANSAC, OSAC and 2SAC-GC methods. Fig. 6 gives the results of setting this parameter. We can find that the ratios of correct registration are obviously improved as the number of iteration increased from 100 to 300, and then improved in a slight manner as the iteration counts further increased. Based on this observation together with a trade-off between efficiency and robustness, the iteration count is consistently set to 300 in this paper. In addition, considering that 300 is larger than the number of correspondences inputted in 1P-RANSAC (i.e., 150), 1P-RANSAC is performed for traversing all 150 inputted correspondences and not implemented in an iterative way.

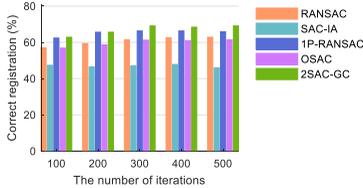

Fig. 6. The ratios of correct registration with repect to different number of iteration.

### 3.4. Test Terms

#### 3.4.1. Performance on Descriptor Based Correspondences

*(1) Performance on different datasets*: In order to comprehensively evaluating the performance, we test the eleven methods on all the four datasets (i.e., S3M, U3M, K3R and S3R datasets) listed in Section 3.1.1. As reported in Table 2, the datasets are acquired by Minolta vivid, Microsoft Kinect and SpaceTime Stereo, respectively. These acquisitions are commonly used in practice. Thus, this evaluation can reflect the performance of these methods in practical application. In addition, the number of matching pairs contained in these four datasets (2339 pairs in total) is abundant, which further ensure an unbiased evaluation for these methods.

*(2) Robustness on varying overlap ratios*: In practice, the scans are usually acquired from different viewpoints, and the overlap ratios of these scans are different. The overlap ratios will influence the percentages of correct correspondences (PCC), and thus impact the performance of transformation estimation. In general, a high overlap ratio can lead to a superior performance for transformation estimation. In order to present the influence of varying overlap ratios to the performance of these transformation estimation techniques, the ratios of correct registration under varying overlap ratios are tested in this paper.

In this test, the overlap ratios are divided into 9 groups which are the ranges increased from 10% to 100% with an interval of 10%. For each transformation estimation technique, the ratio of correct registration is counted with respect to each of the groups. For simplicity, this evaluation is only performed on U3M dataset.

*(3) Combination with the iterative closest point (ICP)*: A standard pipeline of pairwise registration includes coarse and fine registration. All the tested transformation estimation techniques only can coarsely align two scans. In practice, the results of coarse registration are usually refined by fine registration algorithm. Thus, the performance evaluation for the combination of these transformation estimation techniques with a fine registration method presents more practical significance.

The iterative closest point (ICP) [25, 50], introduced in the early 1990s, is the most well-known and popular algorithm for efficiently registering two point clouds under rigid transformation. In this test, we select ICP algorithm to implement fine registration. For comprehensive evaluation, we also use all the four datasets in this test.

*(4) Combination with different local descriptors*: The correspondences generated by different local descriptors commonly have different PCC, and thus these generated correspondences usually exhibit different impacts to the performance of the transformation estimation methods. In this evaluation, we select five state-of-the-art local descriptors (including SHOT [51], RoPS [9], TOLDI [44], MaSH [11] and SDASS [45]) to build correspondences, and then use these correspondences to test all the transformation estimation methods. For simplicity, this test is only performed on U3M dataset.

*(5) Combination with different LRF/A*: In all the eleven tested techniques, CCV, 1P-RANSAC and LGV are performed on the basis of an LRF, and 2SAC-GC is implemented on the basis of an LRA. Clearly, the performance of the above four methods are influenced by the repeatability of LRF/A. Thus, the test of these techniques on different LRF/A is helpful for comprehensively evaluating their performance.

In this evaluation, five state-of-the-art LRF techniques are used for combining the four transformation estimation methods, i.e., CCV, 1P-RANSAC, LGV and 2SAC-GC. These LRF techniques are proposed by Mian et al. [6], Tombari et al. [51], Guo et al. [9], Petrelli et al. [37] and Yang et al. [44], respectively. For fair comparison between LRF and LRA based methods, LRA is defined as the z axis of the LRF used currently. For simplicity, this evaluation is only performed on U3M dataset.

#### 3.4.2. Performance on Synthetic Correspondences

In contrast to descriptor based correspondences, synthetic correspondences are only influenced by the value of PCC, and not affected by the nuisances (e.g., noise, occlusion, symmetrical surface and etc.) contained in datasets. In this test, the correspondences are synthesized with different PCC. PCC value directly impacts the correctness of an estimated transformation. Clearly, a high PCC can easy produce a correct transformation estimation, which generally does not have challenge for a transformation estimation technique. A real challenge for a transformation estimation technique is to estimate the correct transformation between two scans with a low PCC value. For comprehensively evaluating these methods, ten levels of PCC value, increased from 5% to 50% with an interval of 5%, are considered. In general, 5% is a very challenging PCC value and 50% is a prosperous one to a transformation estimation technique. Fig. 7 illustrates the cases with 5% and 50% PCC values. Since the synthetic PCC is not influenced by the variation of datasets, this evaluation is only performed on U3M dataset.

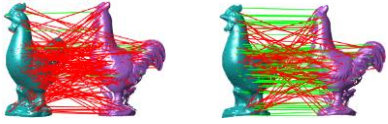

Fig. 7. An illustration of two group of correspondences with 5% (left) and 50% (right) PCC values respectively.

### 3.4.3. Efficiency

Efficiency is also an important attribute for a transformation estimation technique, especially in some time-crucial applications (e.g., robotics and mobile phones). Thus, an evaluation for the computational times of these techniques is also performed. In this evaluation, we count the computational times of these transformation estimation methods on all the four datasets, and the average time costs of these methods for registering one valid scan pair are evaluated in Section 4.3.

## 4 EXPERIMENTAL RESULTS

This section gives the results of the test terms introduced in Section 3.4 together with necessary discussions and explanations.

### 4.1. Results on Descriptor Based Correspondences

#### 4.1.1. Performance on Different Datasets

Fig. 8 gives the results of all methods tested on the four datasets (i.e., S3M, U3M, K3R and S3R datasets). On S3M dataset, V-GTM, 1P-RANSAC and 2SAC-GC are the three best methods, and their performance are very close and obviously outperform the others. The reason for the superiority of these three methods are presented as follows: first, the performance of V-GTM is not influenced by the repeatability of LRF/A; second, 1P-RANSAC can traverse all input correspondences for finding correct ones; third, 2SAC-GC is performed on the basis of an LRA which has higher repeatability than the corresponded LRF [37]. Besides, the performance of GTM is also superior, while its performance has an obvious gap to the performance of V-GTM. The reason is that a geometrical constraint, calculated by the formula (4), used in V-GTM reduces the compatibility between two correspondences with a large geometric deviation. GCM and GCC are the two worst methods, especially for GCC whose performance is obviously inferior to the others. It is mainly caused by that, contrast to GTM and V-GTM methods using iterative evolution to refine consensus set, these two methods directly use the consensus set extracted by the formula (1). In terms of the results on U3M dataset, it can be observed that V-GTM, 1P-RANSAC and 2SAC-GC achieve the best performance and GCC exhibits the worst performance, which are the same with the results on S3M dataset. LGV, OSAC, GTM and RANSAC also achieve superior performance, and their performance are very close. In contrast to the performance on S3M dataset, CVV and SAC-IA have low relative scores on this dataset. For the results on K3R dataset, the performance of all methods is obviously inferior to that on the S3M and U3M datasets, which is caused by the low-quality data obtained with Kinect sensor. We can also observe that V-GTM, GTM and 1P-RANSAC achieve the top three best performance. Compared to the results on S3M and U3M datasets, the rankings of GTM and SAC-IA have an obvious improvement while the ranking of LGV has a clear drop. LGV, CCV, GCM and GCC exhibit inferior performance, and their performance are worse than the others by a large margin. In items of the results on S3R dataset which is also a low-quality dataset with noise and outliers, the overall performance of all methods is very poor, and has a slight drop compared with that on the K3R dataset. Consequently, the differences among the performance of all tested methods are very small. Among these methods, V-GTM, 1P-RANSAC, 2SAC-GC and GTM achieve a relatively better performance, and CCV, GCM and GCC exhibit a relatively worse performance.

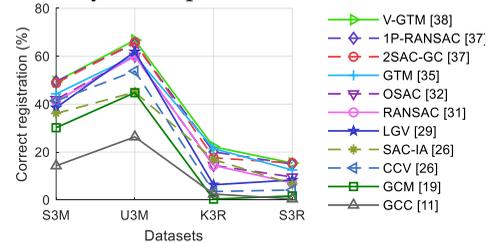

Fig. 8. The percentages of correct registration for the eleven transformation estimation techniques evaluated on the four datasets. The numbers in square brackets are the average percentages computed over the corresponding curves, and listed in a descending order.

For the overall performance of these techniques tested on all four datasets, several conclusions can be concluded as follows. *First*, V-GTM, 1P-RANSAC and 2SAC-GC are the three best techniques. GCC and GCM are the two worst techniques, and the overall performance of them are significantly inferior to the others. It is because that these two methods directly extract maximum consensus set by using the rigidity constraint presented in the formula (1), and do not further refine them. *Second*, V-GTM, 1P-RANSAC, 2SAC-GC, GTM, OSAC and RANSAC generally exhibit a more stable performance across all four datasets compared to the others. In contrary, the performance of LGV, SAC-IA, CCV and GCM varies significantly. *Third*, the overall performance of all methods on K3R and S3R datasets is obviously inferior to that on S3M and U3M datasets. It is because that high noise in these two datasets results in low PCC values and poor repeatability of LRF/A.

#### 4.1.2. Robustness to Overlap Ratios

Fig. 9 presents the percentages of correct registration with respect to vary overlap ratios tested on the U3M dataset. We can observe that the correct percentages of all evaluated techniques are almost consistently increased along with the improvement of overlap ratios. It is clear that the performance of all tested transformation estimation techniques is positive correlation to the value of overlap ratio. All the methods, except for CCV, GCC, GCM and LGV, achieve almost 100% correct registration under the overlap ratios in [60%, 100%]. CCV, SAC-IA and GCM achieve almost 100% correct registration under the overlap ratios in [70%, 100%], [80%, 100%] and [90%, 100%], respectively. The performance of all the methods, except for SAC-IA, GCM and GCC, has a significant improve-ment under overlap ratios increased from 20% to 50%. All methods are almost failure under the overlap ratios in [10%, 20%). GCC is the most sensitive method to small overlap ratios, and almost failure under the overlap ratios below 50%. The worst robustness for GCC is caused by that the clustering directly formed by formula (1) is very vulnerable to low PCC values.

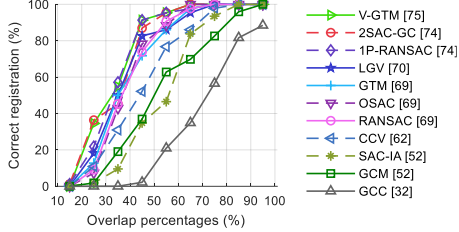

Fig. 9. The percentages of correct registration for the eleven transformation estimation techniques evaluated on the U3M dataset with respect to different overlap percentages. The numbers in square brackets are the average percentages computed over the corresponding curves, and listed in a descending order.

### 4.1.3. Combination with ICP Algorithm

The results of these methods combined with ICP are shown in Fig. 10. We can clearly observe that the overall performance of these transformation estimation methods is obviously improved compared to the results of these methods being not combined with ICP as reported in Section 4.1.1. This is caused by that ICP can further improve the accuracy of the registered results. On S3M dataset, we can observe that the performance of 1P-RANSAC, GTM and OSAC has more obvious improvements than the others after being combined with ICP, which leads to that 1P-RANSAC achieves the best performance and the performance of GTM is comparable to that of 2SAC-GC and V-GTM. We can also observe that GCM and GCC present the worst performance in this test, which are similar to the results of them not being combined with ICP. For the results on U3M dataset, V-GTM achieves the best performance. 2SAC-GC, 1P-RANSAC, GTM, RANSAC and LGV exhibit a good performance, and the gaps among them are small. Similar to the results on S3M dataset, GCC and GCM also have the worst performance on this dataset. Compared to the results of not being combined with ICP, the improvements of V-GTM and SAC-IA are more obvious than the others. On the low-quality datasets (i.e., K3R and S3R), 1P-RANSAC exhibits the best performance. Other good methods include GTM, V-GTM, 2SAC-GC, SAC-IA and OSAC. LGV and RANSAC exhibit mediocre performance. CCV, GCM and GCC are the three worst methods on the low-quality datasets, especially for GCM which is almost failure on the both low-quality datasets. It is clear that GCM, GCC, CCV, RANSAC and LGV are sensitive to low-quality datasets. Compared to the results of not being combined with ICP, the gaps among the performance of all tested methods are larger.

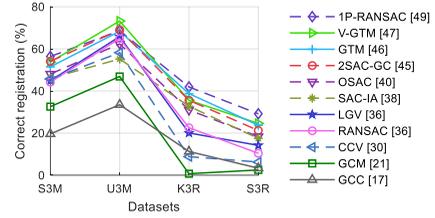

Fig. 10. The performance of the eleven transformation estimation techniques combined with ICP evaluated on the four datasets. The numbers in square brackets are the average percentages computed over the corresponding curves, and listed in a descending order.

### 4.1.4. Combination with Different Local Features

The performance of these methods combined with five local descriptor features (i.e., SHOT [51], RoPS [9], TOLDI [44], MaSH [11] and SDASS [45]) are shown in Fig. 11. We can observe that the best performance of all tested transformation estimation methods is achieved when they combined with SDASS descriptor, which is mainly caused by the superior descriptiveness and robustness of SDASS [45]. We can further observe that 1P-RANSAC achieves the best performance and outperforms the others by a large margin when combined with SHOT, RoPS and TOLDI descriptors. It is because that the features generated by the three descriptors (i.e., SHOT, RoPS and TOLDI) are all based on LRF, which enables that the correct correspondences usually correspond to the LRFs with high repeatability. Thus, if the correspondences built by LRF-based descriptors (e.g., the above three descriptors), 1P-RANSAC is the most appropriate transformation estimation technique. Besides, it can be also observed that the rankings of these methods combined with the five local descriptors are similar. Specifically, on all five cases of this test, 1P-RANSAC, 2SAC-GC and V-GTM are the three best methods. LGV, RANSAC, OSAC, CCV and GTM present a medium performance, and SAC-IA, GCM and GCC have the three worst performance.

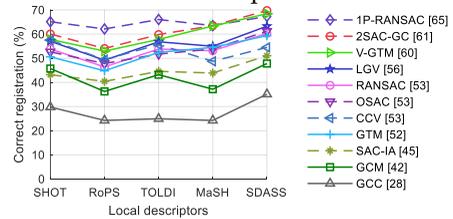

Fig. 11. The percentages of correct registration for the eleven transformation estimation techniques evaluated on U3M dataset with respect to different local features. The numbers in square brackets are the average percentages computed over the corresponding curves, and listed in a descending order.

### 4.1.5. Combination with Different LRF/A

The results of this test are presented in Fig. 12. In addition, to present the relativity between the repeatability of an LRF and the performance of a transformation estimation technique, the repeatability of the five LRF techniques (i.e., Main's LRF [6], Tombari's LRF [51], Guo's LRF [9], Petrelli's LRF [37] and Yang's LRF [44]) is also tested. We can observe that the performance of the four LRF/A based methods (i.e., LGV, 1P-RANSAC, 2SAC-GC and CCV) is positive correlation to the repeatability of the LRF techniques. Thus, it is clear that a high repeatability for an LRF technique can improve the performance of these transformation estimation methods. We can also observe that 2SAC-GC achieves the best overall performance, and its performance outperforms the others by a large margin when combined with Main's and Tombari's LRFs. It is mainly because that 2SAC-GC is only influenced by the error of Z axis in an LRF, while the others are affected by the errors of both the X and Z axis in an LRF. 1P-RANSAC and LGV achieve medium, and CCV exhibits the worst performance.

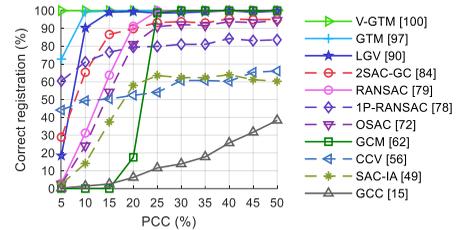

Fig. 13. The ratios of correct registration for the eleven transformation estimation techniques evaluated on synthetic correspondences with respect to different PCC. The numbers in square brackets are the average percentages computed over the corresponding curves, and listed in a descending order.

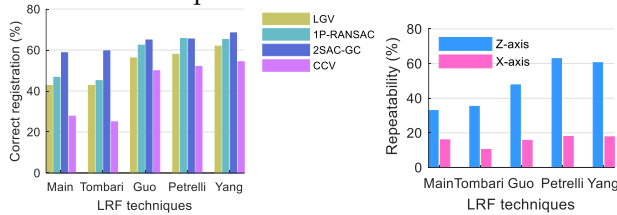

Fig. 12. The performance evaluation of transformation estimation and LRF techniques on U3M dataset. Left: the ratios of correct registration under different LRF techniques. Rgith: the repeatability of different LRF techniques.

### 4.2. Results on Synthetic Correspondences

Fig. 13 gives the performance of all evaluated methods tested on synthetic correspondences with respect to different PCC values. We find that V-GTM achieves 100% correct registration under all PCC cases, and GTM achieves 100% correct registration under the PCC value larger than 5%. These two methods achieve the best performance, and outperform the others by a large margin. In contrast to the results tested on descriptor based correspondences as presented in Section 4.1.1, the performance of V-GTM and GTM have obvious improvement in this test. It is mainly because that, in the descriptor based correspondences, the performance of V-GTM and GTM is vulnerable to the false correspondences caused by symmetric surface as illustrated in Fig. 14, while false correspondences in a symmetric manner are generally not existed in synthetic correspondences owing to that these synthetic correspondences are randomly extracted. When the PCC at 5%, RANSAC, OSAC, SAC-IA, GCM and GCC are almost failure, and LGV and 2SAC-GC also exhibit inferior performance. It indicates that the above seven methods have weak robustness to low PCC values. When the PCC larger than 20%, LGV, RANSAC and GCM also achieve 100% correct registration. Thus, LGV, RANSAC and GCM also have superior performance on high PCC, besides V-GTM and GTM. We can also observe that, compared to the performance tested on descriptor based correspondences, the rankings of 2SAC-GC and 1P-RANSAC significantly drop in this test. It is mainly because that, unlike descriptor based correspondences, the correct synthetic correspondences cannot ensure high repeatability for the LRF/A associated with them, especially for the scan pairs with small overlap ratios. The overall performance of GCC, SAC-IA and CCV is the worst, especially for GCC method whose performance is inferior to the others by a large margin.

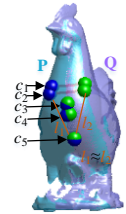

Fig. 14. An illustration of symmetric surface to influence the performance of V-GTM and GTM. $c_1$ and $c_2$ are two false correspondences caused by symmetric surface, and $c_3$, $c_4$ and $c_5$ are three correct correspondences. We can observe that the false correspondences $c_1$ and $c_2$ are compatible with the correct correspondences $c_3$, $c_4$ and $c_5$ tested by the formula (2) and (4), which leads to a incorrect registration since all the five correspondences are regarding as inliers.

### 4.3. Efficiency

The average time costs for the eleven transformation estimation techniques tested on all the four datasets are shown in Fig. 15. These tests are implemented in MATLAB on a PC with 3.6GHz Processor and 12 GB of RAM, and are performed with a single thread.

We can observe that GCC and GCM are the two most efficient methods, and their efficiencies outperform the others by a large margin. It is because GCC and GCM employ less input correspondences. Note that, although they have very high efficiency, the percentages of correct registration achieved by them are very low. RANSAC and LGV are also very efficient methods. GTM and V-GTM achieve a medium efficiency. Note that the above mentioned six methods are all MC-based techniques, and it is clear observed that the efficiency of MC-based techniques is obviously higher than that of CV-based techniques. It is caused by that the CV-based methods need to verify the confidence of a plausible transformation in each iteration. For the efficiency of the CV-based methods, 2SAC-GC, 1P-RANSAC and CCV achieve an acceptable efficiency, while SAC-IA and OSAC are the most time-consuming techniques. The reasons for the superior efficiency of 2SAC-GC, 1P-RANSAC and CCV among CV-based methods are that 2SAC-GC skips some invalid iterations by employing the constraints calculated in the formula (14), and 1P-RANSAC and CCV have less iteration counts (the iteration counts for 1P-RANSAC and CCV being 150 and 200, respectively).

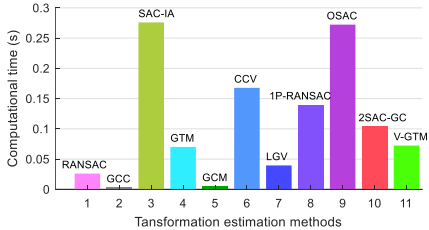

Fig. 15. The efficiencies of the eleven transformation estimation techniques evaluated on U3M dataset.

## 5 SUMMARY AND DISCUSSION

Based on the tested results, the summary and discussion are given in this section. Specifically, we first give the performance summary of all evaluated methods, and then a guidance is provided for selecting an appropriate method in a specific application.

### 5.1. Performance Summary

Taking altogether, the overall performance of V-GTM is the best. It is because that this method not only achieves the highest ratio of correct registration but also is very efficient. In addition, 1P-RANSAC and 2SAC-GC are also superior methods. The performance of the above three methods are very stable across all the datasets. In opposite to the performance of the above three methods, GCC and GCM are the two worst methods owing to that they consistently have the lowest ratio of correct registration on all the datasets though they are very efficient.

Based on the evaluated results, six specific finds are summarized as follows. (i) Using iterative evolution with a game theory manner can effectively exclude false correspondences. It can be clearly verified by that V-GTM and GTM have obviously higher ratios of correct registration than GCC and GCM which only use the geometric constraint as calculated in formula (1) to extract inliers and do not perform iteratively evolution. (ii) The correspondences generated by LRF-based descriptors (e.g., RoPS, SHOT and TOLDI) are beneficial to the performance of LRF-based transformation estimation techniques (e.g., 1P-RANSAC). For instance, the performance of 1P-RANSAC tested on the correspondences generated by these LRF-based descriptors are obviously superior to the others as verified in Section 4.1.4. (iii) Adding the geometric constraint as calculated in formula (1) into the payoff function of GTM with an appropriate way can effectively improve its performance, e.g., V-GTM is constructed by adding the geometric constraint with an exponential form into the payoff function of GTM, and the overall performance of V-GTM is obviously superior to that of GTM. (iv) In CV-based methods, there are three popular techniques (i.e., using three points, LRA and LRF) for determining a plausible transformation. For the comparison among these three techniques based methods, the performance of LRF based technique (e.g., 1P-RANSAC) is similar with that of LRA based technique (e.g., 2SAC-GC), and their performance are obviously superior to the performance of three points based techniques (e.g., RANSAC, SAC-IA and OSAC) when selecting an appropriate LRF technique, such as Yang's LRF technique [44]. (v) Counting the number of inliers in the CV-based methods is the most effective method for verifying the confidence of a transformation so far. For instance, 1P-RANSAC, 2SAC-GC, using the number of inliers to verify confidence, has obviously superior to SAC-IA and OSAC which employ the formulas (12) and (13) to verify confidence, respectively. (vi) The efficiency of MC-based methods is generally higher than that of RV-based methods. It is mainly caused by that RV-based methods need to verify the confidence of a plausible transformation in each iteration.

### 5.2. Application Guidance

For quickly selecting an appropriate transformation estimation method in a particular application, several points of guidance are summarized as follows. (i) On high-quality datasets, V-GTM is the best selection. Although the performance of 1P-RANSAC and 2SAC-GC is comparable with that of V-GTM in terms of correct registrations, V-GTM is more efficient than 1P-RANSAC and 2SAC-GC. (ii) On low-quality dataset, 1P-RANSAC is the best selection owing to that it achieves the highest percentage of correct registration after combining with ICP. In addition, V-GTM, GTM and 2SAC-GC are also good options. (iii) For the applications with the correspondences generated by LRF-based descriptors, 1P-RANSAC is the best option, since this technique has obvious superiority to the others when evaluated on these correspondences. (iv) For time-crucial applications (e.g., robotics and mobile platforms), RANSAC is the best option because that it is very efficient and also achieve acceptable performance in terms of the percentage of correct registration. Besides, V-GTM is also a suitable option since it not only has the best overall performance for the percentage of correct registration but also exhibits a superior execution speed. Note again that although GCC and GCM are more efficient than RANSAC and V-GTM, they suffer from the lowest percentage of correct registration on the most datasets.

## 6 CONCLUSION

In this paper, a comprehensive evaluation of the eleven transformation estimation techniques has been conducted on four benchmark datasets acquired with different devices (including Minolta vivid, Kinect and Space Time). Specifically, the ratios of correct registration achieved by these techniques were evaluated on descriptor based and synthetic correspondences, respectively. On descriptor based correspondences, several items (including performance on different datasets, robustness to different overlap ratios, combination with ICP, different local descriptors and different LRF/A techniques) were evaluated. On synthesis correspondences, the robustness of these methods to different PCC values are evaluated. In addition, the computational efficiencies of these transformation estimation techniques were also evaluated. Finally, the merits and demerits of these methods were summarized and discussed, and the application guidance were also provided. To the best of our knowledge, this is the first work of comprehensively evaluating the transfor-mation estimation techniques in 3D space. This paper can be regarded as a guide for users to select the most appropriate transformation estimation method.

## ACKNOWLEDGMENT

This work was jointly supported by the national natural science foundation of china under Grant 51575354, the national science and technology major project under Grant 2017ZX04016001 and the shanghai municipal science and technology project under Grant 18511107302.

## REFERENCES

[1] J. Yang, Z. Cao and Q. Zhang, A fast and robust local descriptor for 3D point cloud registration, *Inform Sciences*, vol. 346-347, pp. 163-179, 2016.

[2] Y. Guo, F. Sohel and M. Bennamoun, et al., An Accurate and Robust Range Image Registration Algorithm for 3D Object Modeling, *Ieee T Multimedia*, vol. 16, no. 5, pp. 1377-1390, 2014.

[3] A. Aldoma, F. Tombari and L.D. Stefano, et al., A Global Hypothesis Verification Framework for 3D Object Recognition in Clutter, *Ieee T Pattern Anal*, vol. 38, no. 7, pp. 1383-1396, 2016.

[4] E. Rodolà, A. Albarelli and F. Bergamasco, et al., A Scale Independent Selection Process for 3D Object Recognition in Cluttered Scenes, *Int J Comput Vision*, vol. 102, no. 1-3, pp. 129-145, 2013.

[5] A.S. Mian, M. Bennamoun and R. Owens, Three-Dimensional Model-Based Object Recognition and Segmentation in Cluttered Scenes, *Ieee T Pattern Anal*, vol. 28, no. 10, pp. 1581-1601, 2006.

[6] A. Mian, M. Bennamoun and R. Owens, On the Repeatability and Quality of Keypoints for Local Feature-based 3D Object Retrieval from Cluttered Scenes, *Int J Comput Vision*, vol. 89, no. 2-3, pp. 348-361, 2010.

[7] A.M. Bronstein, M.M. Bronstein and L.J. Guibas, et al., Shape Google: Geometric Words and Expressions for Invariant Shape Retrieval, *Acm T Graphic*, vol. 30, no. 1, pp. 1-20, 2011.

[8] Y. Gao and Q. Dai, View-Based 3D Object Retrieval: Challenges and Approaches, *Ieee Multimedia*, vol. 21, no. 3, pp. 52-57, 2014.

[9] Y. Guo, F. Sohel and M. Bennamoun, et al., Rotational Projection Statistics for 3D Local Surface Description and Object Recognition, *Int J Comput Vision*, vol. 105, no. 1, pp. 63-86, 2013.

[10] Y. Guo, F. Sohel and M. Bennamoun, et al., A novel local surface feature for 3D object recognition under clutter and occlusion, *Inform Sciences*, vol. 293, pp. 196-213, 2015.

[11] J. Yang, Q. Zhang and Z. Cao, Multi-attribute statistics histograms for accurate and robust pairwise registration of range images, *Neurocomputing*, vol. 251, pp. 54-67, 2017.

[12] Y. Guo, B. Mohammed and S. Ferdous, et al., 3D Object Recognition in Cluttered Scenes with Local Surface Features: A Survey, *Ieee T Pattern Anal*, vol. 36, no. 11, 2014.

[13] Y. Guo, M. Bennamoun and F. Sohel, et al., A Comprehensive Performance Evaluation of 3D Local Feature Descriptors, *Int J Comput Vision*, vol. 116, no. 1, pp. 66-89, 2016.

[14] H. Lei, G. Jiang and L. Quan, Fast Descriptors and Correspondence Propagation for Robust Global Point Cloud Registration, *Ieee T Image Process*, pp. 1, 2017.

[15] M.A. Fischler and R.C. Bolles, Random Sample Consensus: A Paradigm for Model Fitting with Applicatlon to Image Analysis and Automated Cartography, *Commun Acm*, vol. 24, no. 6, pp. 381-395, 1981.

[16] R.B. Rusu, N. Blodow and M. Beetz, Fast Point Feature Histograms (FPFH) for 3D Registration, *Proc. IEEE International Conference on Robotics and Automation*, 2009, pp. 3212-3217.

[17] A. Albarelli, S.R. Bulò and A. Torsello, et al., Matching as a non-cooperative game, *Proc. IEEE International Conference on Computer Vision*, 2009, pp. 1319-1326.

[18] A. Albarelli, E. Rodolà and A. Torsello, Fast and accurate surface alignment through an isometry-enforcing game, *Pattern Recogn*, vol. 48, no. 7, pp. 2209-2226, 2015.

[19] Y. Liu, L. De Dominicis and B. Wei, et al., Regularization Based Iterative Point Match Weighting for Accurate Rigid Transformation Estimation, *Ieee T Vis Comput Gr*, vol. 21, no. 9, pp. 1058-1071, 2015.

[20] F. Tombari, S. Salti and L. Di Stefano, Performance Evaluation of 3D Keypoint Detectors, *Int J Comput Vision*, vol. 102, no. 1-3, pp. 198-220, 2013.

[21] J. Yang, Y. Xiao and Z. Cao, Toward the Repeatability and Robustness of the Local Reference Frame for 3D Shape Matching: An Evaluation., *Ieee T Image Process*, vol. PP, no. 99, pp. 1, 2018.

[22] Y. Díez, F. Roure and X. Lladó, et al., A Qualitative Review on 3D Coarse Registration Methods, *Acm Comput Surv*, vol. 47, no. 3, pp. 1-36, 2015.

[23] G.K.L. Tam, Z. Cheng and Y. Lai, et al., Registration of 3D Point Clouds and Meshes: A Survey from Rigid to Nonrigid, *Ieee T Vis Comput Gr*, vol. 19, no. 7, pp. 1199-1217, 2013.

[24] J. Salvi, C. Matabosch and D. Fofi, et al., A review of recent range image registration methods with accuracy evaluation, *Image Vision Comput*, vol. 25, no. 5, pp. 578-596, 2007.

[25] P.J. Besl and N.D. McKay, A method for registration of 3-D Shapes, *IEEE Transaction on pattern analysis and machine intelligence*, vol. 14, no. 2, pp. 239-256, 1992.

[26] H. Chen and B. Bhanu, 3D free-form object recognition in range images using local surface patches, *Pattern Recogn Lett*, vol. 28, no. 10, pp. 1252-1262, 2007.

[27] D. Thomas and A. Sugimoto, Robustly registering range images using local distribution of albedo, *Comput Vis Image Und*, vol. 115, no. 5, pp. 649-667, 2011.

[28] A. Albarelli, E. Rodola and A. Torsello, A game-theoretic approach to fine surface registration without initial motion estimation, *Proc. Computer Vision and Pattern Recognition*, 2010, pp. 430-437.

[29] S. Umeyama, least-squares estimation of transformation parameters between two point patterns, *Ieee T Pattern Anal*, vol. 13, no. 4, pp. 376-380, 1991.

[30] B.K. Horn, H.H. M and S. Negahdaripour, Closed-form solution of absolute orientation using orthonormal matrices, *Journal of the Optical Society of America A: Optics and Image Science, and Vision*, vol. 5, no. 7, pp. 1127-1135, 1988.

[31] B.K.P. Horn, Closed-form solution of absolute orientation using unit quaternions, *Journal of the Optical Society of America A: Optics and Image Science, and Vision*, vol. 4, no. 4, pp. 629-642, 1987.

[32] D. Zai, J. Li and Y. Guo, et al., Pairwise registration of TLS point clouds using covariance descriptors and a non-cooperative game, *Isprs J Photogramm*, vol. 134, pp. 15-29, 2017.

[33] A.G. Buch, Y. Yang and N. Kruger, et al., In Search of Inliers: 3D Correspondence by Local and Global Voting, *Proc. IEEE Conference on Computer Vision and Pattern Recognition*, 2014.

[34] N. Ohtsu, A Threshold Selection Method from Gray-Level Histograms, *IEEE Transactions on Systems Man & Cybernetics*, vol. 9, no. 1, pp. 62-66, 2007.

[35] Y. Guo, B. Mohammed and S. Ferdous, et al., An Integrated Framework for 3-D Modeling, Object Detection, and Pose Estimation From Point-Clouds, *Ieee T Instrum Meas*, vol. 60, no. 3, 2015.

[36] J. Yang, Q. Zhang and Z. Cao, The effect of spatial information

characterization on 3D local feature descriptors: A quantitative evaluation, *Pattern Recogn*, vol. 66, pp. 375-391, 2017.
[37] A. Petrelli and L. Di Stefano, On the repeatability of the local reference frame for partial shape matching, *Proc. 2011 International Conference on Computer Vision*, 2011, pp. 2244-2251.
[38] B. Curless and M. Levoy, A Volumetric Method for Building Complex Models from Range Images, *Proc. 23rd Annual Conference on Computer Graphics and Interactive Techniques*, 1996, pp. 303-312.
[39] A.S. Mian, M. Bennamoun and R.A. Owens, A Novel Representation and Feature Matching Algorithm for Automatic Pairwise Registration of Range Images, *Int J Comput Vision*, vol. 66, no. 1, pp. 19-40, 2006.
[40] A. Tonioni, S. Salti and F. Tombari, et al., Learning to Detect Good 3D Keypoints, *Int J Comput Vision*, 2017.
[41] M. Brown, D. Windridge and J. Guillemaut, A generalised framework for saliency-based point feature detection, *Comput Vis Image Und*, vol. 157, pp. 117-137, 2017.
[42] Y. Liu, R.R. Martin and L. de Dominicis, et al., Using retinex for point selection in 3D shape registration, *Pattern Recogn*, vol. 47, no. 6, pp. 2126-2142, 2014.
[43] I. Sipiran and B. Bustos, Harris 3D: a robust extension of the Harris operator for interest point detection on 3D meshes, *The Visual Computer*, vol. 27, no. 11, pp. 963-976, 2011.
[44] J. Yang, Q. Zhang and Y. Xiao, et al., TOLDI: An effective and robust approach for 3D local shape description, *Pattern Recogn*, vol. 65, pp. 175-187, 2017.
[45] B. Zhao, X. Le and J. Xi, A Novel SDASS Descriptor for Fully Encoding the Information of 3D Local Surface, *arXiv:1711.05368*, 2017.
[46] S. Salti, F. Tombari and L. Di Stefano, SHOT: Unique signatures of histograms for surface and texture description, *Comput Vis Image Und*, vol. 125, pp. 251-264, 2014.
[47] D.G. Lowe, Distinctive Image Features from Scale-Invariant Keypoints, *Int J Comput Vision*, vol. 60, no. 2, pp. 91-110, 2004.
[48] T. Weber, R. Hänsch and O. Hellwich, Automatic registration of unordered point clouds acquired by Kinect sensors using an overlap heuristic, *Isprs J Photogramm*, vol. 102, pp. 96-109, 2015.
[49] S. Quan, J. Ma and F. Hu, et al., Local voxelized structure for 3D binary feature representation and robust registration of point clouds from low-cost sensors, *Inform Sciences*, 2018.
[50] Y. Chen and G. Medioni, Object Modeling by Registration of Multiple Range Images, *Image Vision Comput*, vol. 10, no. 3, pp. 145-155, 1992.
[51] F. Tombari, S. Salti and L. Di Stefano, Unique Signatures of Histograms for Local Surface Description, *Proc. European Conference on Computer Vision*, Springer, 2010, pp. 356-369.